%% file: main_arxiv.tex
\documentclass{article}

\PassOptionsToPackage{numbers, sort}{natbib}

\usepackage{arxiv}
\usepackage{amsmath}
\usepackage{amsthm}

\usepackage[utf8]{inputenc} %
\usepackage[T1]{fontenc}    %
\usepackage{hyperref}       %
\usepackage{url}            %
\usepackage{booktabs}       %
\usepackage{amsfonts}       %
\usepackage{nicefrac}       %
\usepackage{microtype}      %
\usepackage{xcolor}         %
\input{my_cmd}              %
\input{math_commands}

\definecolor{citecolor}{HTML}{0071BC}
\definecolor{linkcolor}{HTML}{ED1C24}
\hypersetup{colorlinks=true, linkcolor=linkcolor, citecolor=citecolor,urlcolor=black}

\title{%
  \includegraphics[height=\baselineskip]{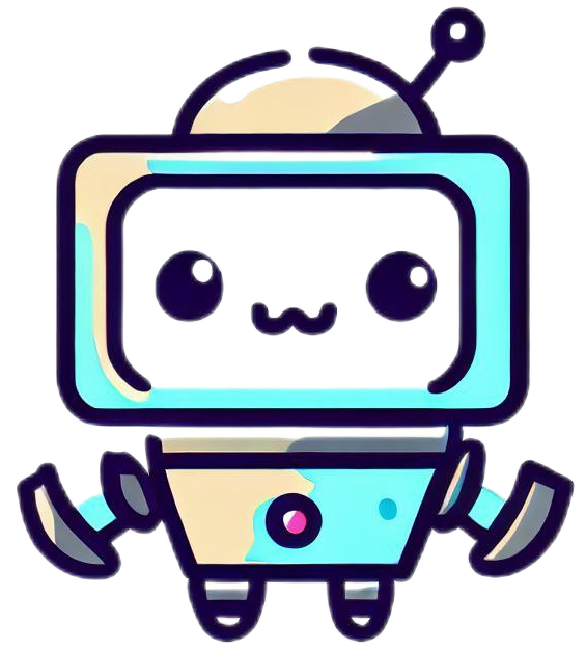}\, MM-LIMA:
  A 200-Instruction Paradigm\\ for Fine-Tuning MiniGPT-4%
}

\author{%
\textbf{Lai Wei$^{1}$
  \quad
  Xiaozhe Li$^{4}$
  \quad
  Zihao Jiang$^{1}$
  \quad
  Weiran Huang$^{1,2}$\thanks{Correspondence to Weiran (weiran.huang@outlook.com).}
  \quad
  Lichao Sun$^3$}\\[0.3cm]
  $^1$School of Computer Science, Shanghai Jiao Tong University \quad $^2$Shanghai Innovation Institute\\[0.1cm]
  $^3$Lehigh University \quad $^4$Tongji University
}

\begin{document}

\maketitle
\begingroup
\renewcommand{\thefootnote}{}%
\footnote{For the officially published version, see \url{https://ietresearch.onlinelibrary.wiley.com/doi/pdf/10.1049/aie2.70001}.}%
\endgroup

\input{sections/abstract}
\input{sections/introduction}

\input{sections/methodology}

\input{sections/experiments}

\input{sections/discussion}

{\bibliography{iclr2024_conference}
 \bibliographystyle{unsrtnat}}

\clearpage
\appendix
\input{sections/appendix}
\end{document}

%% file: my_cmd.tex
\usepackage{multirow}
\usepackage{tabularx}
\usepackage{lipsum} 
\usepackage{graphicx}
\usepackage{makecell}
\usepackage{listings} 
\usepackage{float}
\usepackage{graphicx}
\usepackage{hyperref}
\usepackage{url}
\usepackage{multirow}
\usepackage{tabularx}
\usepackage{lipsum} 
\usepackage{graphicx}
\usepackage{makecell}
\usepackage{listings} 
\usepackage{float}
\usepackage{algorithmic}

\usepackage{url}
\usepackage{multirow}
\usepackage{caption}
\usepackage{subcaption}
\usepackage{algorithm}
\usepackage{bm}
\usepackage{thmtools}
\usepackage{array}
\usepackage{makecell}
\usepackage{amsmath}
\usepackage{booktabs}   
\usepackage{wrapfig}
\usepackage{enumitem}
\usepackage{subcaption}

%% file: math_commands.tex
\usepackage{amsmath,amsfonts,bm}

\def\eqref#1{equation~\ref{#1}}

\def\1{\bm{1}}

\DeclareMathAlphabet{\mathsfit}{\encodingdefault}{\sfdefault}{m}{sl}
\SetMathAlphabet{\mathsfit}{bold}{\encodingdefault}{\sfdefault}{bx}{n}

\DeclareMathOperator*{\argmax}{arg\,max}

%% file: sections/abstract.tex
\begin{abstract}

Multimodal large language models are typically trained in two stages: first pre-training on image-text pairs, and then fine-tuning using supervised vision-language instruction data.
Recent studies have shown that large language models can achieve satisfactory results even with a limited amount of high-quality instruction-following data.
In this paper, we introduce MM-LIMA, which is fine-tuned on a small dataset comprising only 200 examples, amounting to approximately 6\% of the instruction-following data used in the alignment dataset for MiniGPT-4~\cite{zhu2023minigpt}.
To achieve this, we first propose several metrics to access the quality of multimodal instruction data.
Based on these metrics, we present an effective and trainable data selector to automatically identify and filter low-quality vision-language data.
By employing this method, MM-LIMA outperforms the original MiniGPT-4 on various evaluations.
Overall, our findings demonstrate that less but high-quality instruction tuning data is efficient in enabling multimodal large language models to generate better output.
Our code is available at \url{https://github.com/waltonfuture/InstructionGPT-4}.

\end{abstract}

%% file: sections/introduction.tex
\section{Introduction}

GPT-4~\cite{openai2023gpt4} has showcased its powerful prowess in generating highly detailed and precise descriptions of images, signaling a new era of language and visual processing. Thus, GPT-4 like Multimodal Large Language Models (MLLMs) have recently emerged as a prominent research area, harnessing powerful Large Language Models (LLMs) as a cognitive framework for conducting multimodal tasks.
The remarkable and unexpected capabilities exhibited by MLLMs surpass those of traditional methods, indicating a potential pathway towards artificial general intelligence. 
To achieve this, massive image-text pairs and vision-language instruction tuning data have been employed to train simple connectors (e.g., MiniGPT-4~\cite{zhu2023minigpt} and LLaVA~\cite{liu2023llava}) between frozen LLMs (e.g., LLaMA~\cite{touvron2302llama} and Vicuna~\cite{vicuna2023}) and visual representations (e.g., CLIP~\cite{radford2021learning} and BLIP-2~\cite{li2023blip}). 

MLLMs are usually trained in two stages: pre-training and fine-tuning~\cite{zhu2023minigpt, liu2023llava, gao2023llamaadapterv2, dai2023instructblip}.
Pre-training on image-text pairs helps MLLMs gain a large amount of knowledge while fine-tuning teaches models to better understand human intentions and generate accurate responses.
Recently, instruction tuning on large-scale datasets has served as a powerful fine-tuning technique to empower MLLMs with enhanced vision-language understanding and instruction-following abilities~\cite{zhao2023svit, zhang2023llavar, liu2023aligning}. 
It facilitates the alignment of models with human preferences, enabling the generation of desired outputs in response to various instructions.
Recent state-of-the-arts, including InstructBLIP~\cite{dai2023instructblip} and Otter~\cite{li2023otter}, have shown promising results by leveraging a collection of vision-language datasets for visual instruction tuning.

However, it has been observed that commonly used instruction-tuning datasets surprisingly contain numerous low-quality instances with incorrect or irrelevant responses~\cite{zhou2023lima, chen2023alpagasus, cao2023instruction}.
Such data can mislead and negatively impact the performance of the model.
This issue has prompted researchers to delve into the possibility of achieving robust performance using a small quantity of high-quality instruction-following data.
Encouragingly, recent studies have substantiated the promising potential of this approach.
\citet{zhou2023lima} introduce LIMA, a language model fine-tuned with carefully curated high-quality data, selected by human experts.
This study has shown that LLMs can achieve satisfactory results even with a limited amount of high-quality instruction-following data.
The proposed idea ``Less is More'' tells that data quality is more important than data quantity to improve model performance, which does not conflict with the Scaling Law~\cite{kaplan2020scaling}.
Building upon these foundations, our objective is to determine if using less instruction data can yield better alignment results in multimodal large language models.
Nevertheless, there is a challenge that the process of collecting appropriate high-quality vision-language datasets for fine-tuning multimodal language models lacks clear guidelines.

\begin{wrapfigure}{r}{0.4\linewidth}
\vspace{-10pt}
  \centering
    \includegraphics[width=\linewidth]{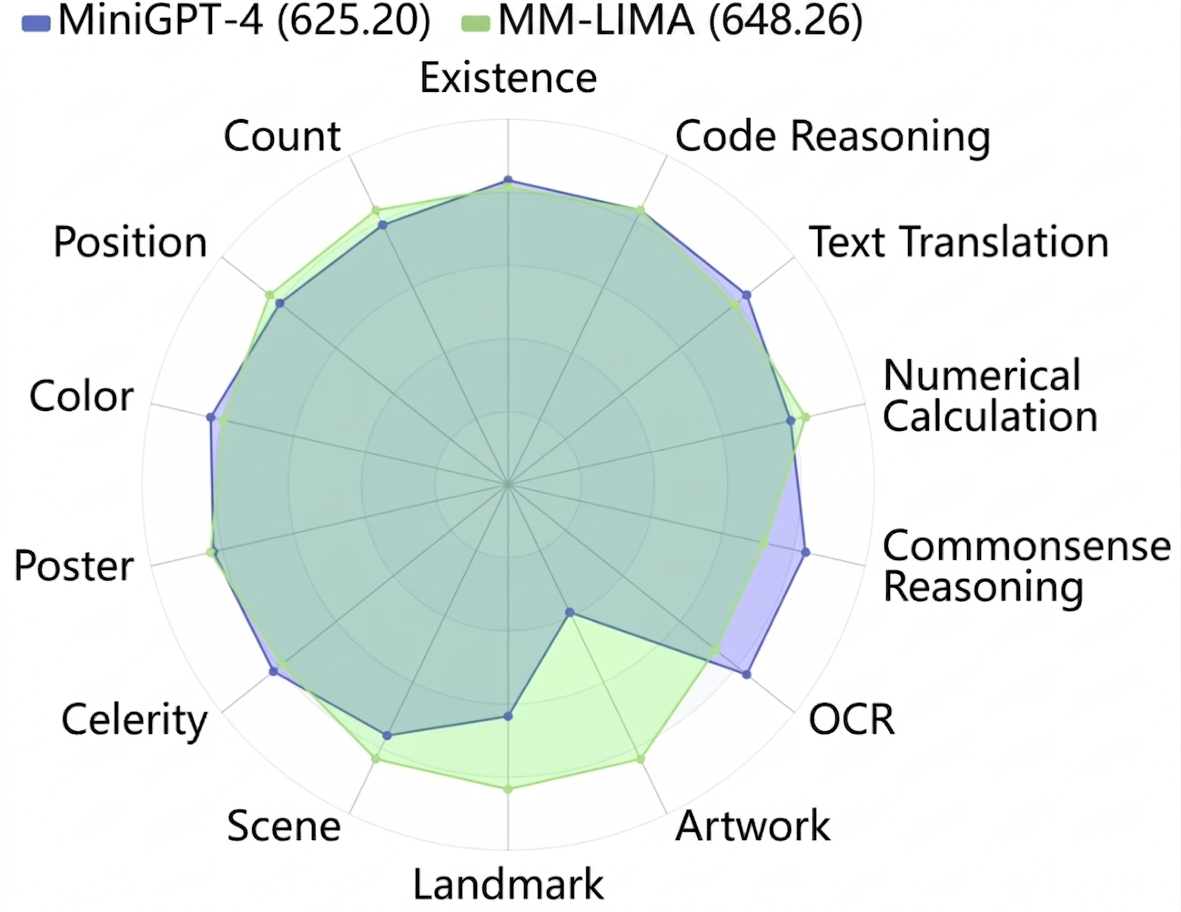} 
  \caption{Comparison of MME evaluation (MM-LIMA vs.\ MiniGPT-4).}\label{fig:radar}
   \vspace{-10pt}
\end{wrapfigure}

Different from LIMA~\cite{zhou2023lima} that requires manually constructed dataset, we aim to propose a robust and effective data selector that automatically identifies and filters low-quality vision-language data from existing datasets, ensuring that our model is trained on the most relevant and informative samples. 
The key focus of our study lies in exploring the efficacy of reduced but high-quality instruction-tuning data in fine-tuning MLLMs.
Another challenge is the current lack of comprehensive methods for evaluating the quality of vision-language data.
Therefore, we introduce several novel metrics tailored for assessing the quality of multimodal instruction data, including CLIP Score~\cite{radford2021learning}, GPT Score~\cite{chen2023alpagasus}, Reward Score~\cite{reward}, Length Score and Multimodal Features of each vision-language data.

To investigate the metrics’ relationship with the real instruction data quality, we first split a range of distinct subsets from the original fine-tuning data. Subsequently, we record each fine-tuned model’s performance on the validation set as the labels of data quality. 
We then compute the metrics and multimodal data features and combine them as an embedding across each subset. 
After that, we apply a self-attention network as the data selector to determine the relationship between the genuine quality labels and embeddings.
We perform spectral clustering~\cite{ng2001spectral}, which aims to ensure the diversity of data distribution, on the original 3.4K data used to fine-tune MiniGPT-4.
Finally, we apply the data selector on each cluster to predict its quality label and sort. 
Through this series of procedures, MM-LIMA is fine-tuned on a much smaller but carefully selected subset following the same training configuration of MiniGPT-4.

Our evaluations focus on a wide range of complex open-domain multimodal large language model benchmarks, including MME~\cite{fu2023mme}, MMBench~\cite{liu2023mmbench}, VQA datasets from LVLM-eHub~\cite{xu2023lvlmehub}, etc.
Through rigorous experimentation, we demonstrate that 200 pieces of data used for fine-tuning, which is 6\% of the original scale, are enough to help MM-LIMA achieve comprehensive superiority over MiniGPT-4 across these diverse multimodal tasks, with a +23 score enhancement on MME, a +1.55 score improvement on MMBench, and a +1.76\% increase in performance on VQA datasets compared to MiniGPT-4. 
Specifically, MM-LIMA outperforms MiniGPT-4 in 8 out of 14 tasks within MME, 13 out of 20 abilities in MMBench, and excels in all four VQA datasets included in LVLM-eHub. 
This discovery is inspiring, as it demonstrates that the data quality in vision-language instruction tuning can outweigh the quantity.
In addition, this shift towards prioritizing data quality presents a new and more efficient paradigm that can generally improve the fine-tuning stage of MLLMs. 

Our contributions are summarized as follows:
\begin{itemize}[itemsep=0pt,topsep=0pt,leftmargin=0.5cm]
  \item We are the first to demonstrate that less instruction data for better alignment is also suitable for multimodal large language model, by showing that fine-tuning MiniGPT-4 with only 6\% instruction-following data can achieve better performance.
  \item We introduce the concept of genuine quality labels along with a set of indicators for evaluating the quality of multimodal instruction-following data, and propose a learnable data selector to obtain high-quality vision-language data for fine-tuning. 
  \item 
  Our MM-LIMA fine-tuned with 200 instructions consistently outperforms the original MiniGPT-4 in various popular benchmarks such as MME, MMBench and VQA datasets.

\end{itemize}

%% file: sections/methodology.tex
\section{MM-LIMA}

\begin{table}[t]
\centering
\renewcommand{\arraystretch}{1.1}
\resizebox{0.9\textwidth}{!}{
\begin{tabular}{c|m{13cm}}
\toprule
{Indicators} & \multicolumn{1}{c}{Explanation} \\
\midrule
{CLIP Score} & {The cosine similarity between image embedding and response text embedding. The CLIP Score serves as a measure of the alignment between the provided image and its accompanying caption. This score quantifies how well the caption accurately describes the visual content, ensuring that the image and text are in concordance.}\\
\midrule
{Length Score} & The length of every answer in the multimodal dataset. The length metric gauges the extent of information encapsulated within the caption. A balanced and informative answer length is crucial to convey the desired instruction without being excessively verbose or overly concise.\\
\midrule
{Reward Score} & Score from a reward model~\cite{reward} that judges the human likeness to a response. The reward model is trained from human feedback to predict which generated answer is better judged by a human, given a question.\\
\midrule
{GPT Score} & Score from GPT4~\cite{openai2023gpt4} to evaluate the quality of response. The GPT Score reflects the LLM's assessment of the caption's quality. This score is indicative of how effectively the generated caption adheres to the model's language proficiency, considering factors such as grammar, semantics, and fluency.\\ 
\midrule
{Multimodal Features} & Vision-language features in low dimensional space obtained by encoding images with ViT from CLIP~\cite{radford2021learning} and text with Llama2~\cite{touvron2023llama}, followed by conducting unsupervised dimensionality reduction.
 \\
\bottomrule
\end{tabular}}
\caption{Quantitative indicators and explanations for evaluating instruction-following data quality. CLIP Score measures the suitability between the image and caption. Length Score, Reward Score, and GPT Score
measure the comprehensive quality of the caption. Multimodal Features represent the essential characteristics of vision-language data. } \label{tab:indicators}
\end{table}

\begin{figure}[t]
\centering
  \includegraphics[width=0.8\textwidth]{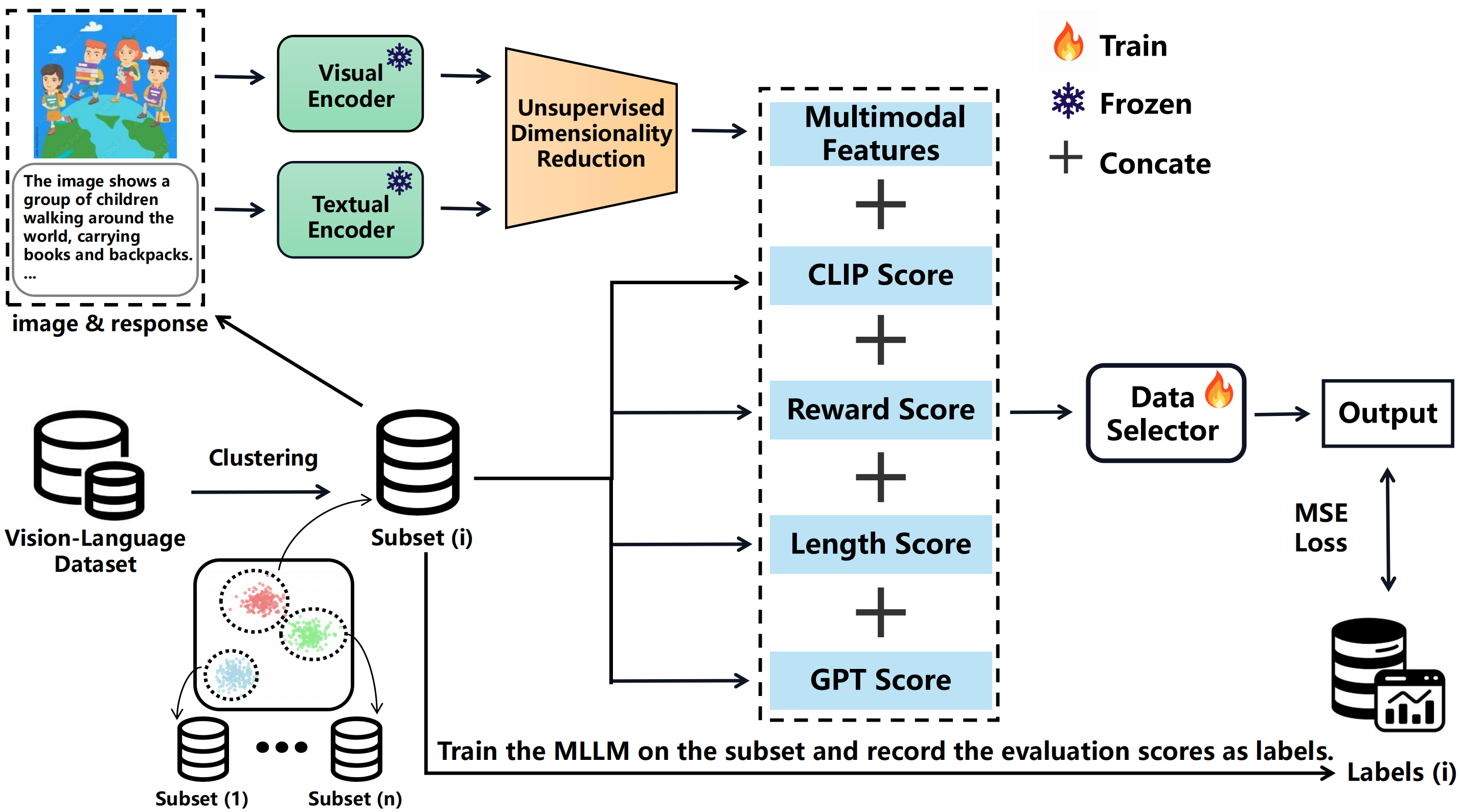} 
  \caption{Overall procedures of the data selector. We first split the vision-language dataset into $n$ subsets. Subsequently, we train the MLLM on each subset and record the evaluation scores as genuine quality labels. Additionally, we concatenate various indicators generated from these subsets to form embeddings. These embeddings were then used to train the data selector, with the objective of aligning the embeddings with the quality labels.} 
  \label{fig:train} 
\end{figure}

In this paper, we present MM-LIMA, a multimodal large language model fine-tuned on a set of 200 high-quality instructions carefully chosen from the dataset utilized in the second-stage training (comprising 3.4K instructions) of MiniGPT-4~\cite{zhu2023minigpt}.
The core of MM-LIMA is the selection of high-quality instructions. 
Thus, in this section, we begin by defining genuine quality labels and presenting indicators for assessing the quality of multimodal instruction-following data. Subsequently, we train a learnable data selector to align these indicators with the genuine labels. An overall procedure of the data selector is illustrated in Figure~\ref{fig:train}.

\subsection{Selecting Principle} \label{principle}
Selecting useful multimodal instruction data is crucial for effectively training MLLMs. Following LIMA~\cite{zhou2023lima}, we propose two key principles for selecting optimal instruction data: diversity and quality. 

\textbf{Diversity.}~~As most of the knowledge is obtained during the pre-training stage for MLLMs, it is necessary to gain better alignment abilities by training on diverse vision-language instruction data.
We adopt spectral clustering on the image embeddings encoded to divide the data into ten categories.
Our ablation study is detailed in Section~\ref{subsection2}.

\textbf{Quality.}~~Vision-language instruction data teaches the multimodal model to follow a certain pattern when interacting with users.
Hence, the quality of these instruction-following data could be viewed as its ability to efficiently steer multimodal language models in learning to generate responses in a particular manner.
In Section~\ref{subsec:qualityeval}, we present our multimodal instruction selection process. We introduce the concept of genuine quality labels along with several related indicators designed for a quantitative assessment of data quality. The specific indicators used for this quantitative evaluation of data quality are outlined in Table~\ref{tab:indicators}, while the genuine quality labels are presented in Table~\ref{tab:qualitylab}.

\subsection{Indicators and Genuine Quality Labels} 
\label{subsec:qualityeval}

Inspired by Instruction Mining~\cite{cao2023instruction} which estimates the data quality by the loss produced by the fine-tuned model, we propose that assessing the real quality of a set of data is contingent on its effectiveness in training a model, i.e., whether the model performs well when trained on this dataset. 
Therefore, we assert that the metrics (e.g., accuracy, F1 score~\cite{sasaki2007truth}) obtained when evaluating the model after training on this dataset can be considered genuine labels for evaluating the quality of this dataset.
However, training an MLLM for evaluation in various datasets can be inefficient. To conveniently assure the quality of the selected multimodal instruction data, we formulate a set of indicators for assessment in Table~\ref{tab:indicators} and train a neural network as a data selector to fit the indicators to the genuine quality labels. 
Thus, the data selector can be applied to other different multimodal datasets directly. 
Here we introduce how to obtain genuine quality labels and indicators for vision-language data.

Given a small set of vision-language instruction data $D$, it is used to fine-tune a pre-trained MLLM, and the fine-tuned model is subsequently evaluated on a series of validation datasets to obtain an average score, which serves as a genuine quality label $y_D$ for the set. 

For the triplets $x$ containing images, instructions, and responses in $D$, we employ CLIP Score~\cite{radford2021learning} $C(x)$ to measure the matching degree between the image and the response.
We also apply the length of responses $L(x)$, and take the Reward Score~\cite{reward} into consideration, which is $R(x)$.
We prompt GPT-4~\cite{openai2023gpt4} as an auto-grader rating each sample $x \in D$ with a GPT Score $G(x, p_G)$ wherein $p_G$ is the rating instruction that is designed based on the prompt from Alpagasus~\cite{chen2023alpagasus} shown in Appendix~\ref{gptprompt}.
These four scores in Table~\ref{tab:indicators} are intuitive and clear, and they can more completely cover the various aspects of multimodal data quality. 
Using a single score to filter data can be useful, but it may not provide a comprehensive measure of data quality. Therefore, it is necessary to combine multiple indicators as an embedding to assess data quality collectively.
By concatenating the four scores with the image-text features, the created embeddings can more comprehensively represent the characteristics of multimodal data. 
The high dimensionality of image-text features, which come from a frozen visual encoder $f$ (e.g., ViT~\cite{ilharco_gabriel_2021_5143773}) and a frozen textual encoder $g$ (e.g., Llama2~\cite{touvron2023llama}), can indeed lead to a large number of parameters that need to be learned during the fitting process, making the task potentially prone to overfitting. Using an unsupervised dimensionality reduction method $P$ (e.g., Principal Component Analysis) that can preserve important information without training is a sensible approach to address this issue.
Consequently, each piece of multimodal data can be assigned an embedding $e$ based on these indicators in Table~\ref{tab:indicators}, i.e.,
\begin{equation}
\label{eq:embed}
e(x) = \mathcal{}{\text{Concat}\, [C(x), L(x), R(x), G(x, p_G), P(f(x_{\text{image}}), g(x_{\text{response}}))]} .
\end{equation}

Our framework is general and not limited to these indicators. Other metrics to measure the quality of multimodal data can also be considered, such as perplexity and Error L2-Norm~\cite{paul2021deep}. We leave exploring possibly more effective and sophisticated architectural designs as future work.

\subsection{Data Selector}
\textbf{Training.}~~Given a vision-language instruction dataset,
 a reasonable and straightforward strategy to obtain the genuine quality labels is to divide the original multimodal dataset into $n$ subsets of equal size through clustering (e.g., $K$-means++).
 For each subset $i$, we now obtain the embedding of every triplet in subset $i$, along with the quality label $y_i$, as detailed in Section~\ref{subsec:qualityeval}. 
We concatenate these embeddings into a single composite embedding denoted as $e_i$, paired with its corresponding quality label $y_i$.
Having gathered a collection of such pairs $(e_i, y_i)$ from all $n$ subsets, we can then proceed to learn a data selector $F$ to fit the embeddings $\{e_i\}_{i=1}^n$ to quality labels $\{y_i\}_{i=1}^n$. The data selector could take various forms, such as a linear layer, an MLP, or a self-attention.

\textbf{Testing.}~~Given a multimodal dataset $D$ of triplets $x$ = (image, instruction, answer) with $x \in D$ and a pre-trained MLLM (e.g., MiniGPT-4), our ultimate objective is to identify a high-quality subset $S \subset D$ that, when utilized for fine-tuning, leads to the improvement of the pre-trained MLLM.

In order to select $S$ from $D$ and ensure its diversity, we first use a clustering algorithm (e.g., spectral clustering) to separate the images in $D$ into $K$ groups.
The clustering algorithm is supposed to be different from the previous one because each of the former clusters shares the same quality label.
Suppose that the total amount of $D$ is $\lvert D \rvert$ and the $i$-th group's amount is $\lvert D_i \rvert$. We set $|S| = \alpha$ as the size of the target subset.

For each $x$ in $D$, we gain an embedding $e(x)$ in Equation~\eqref{eq:embed}.  We sort $x$ according to the predicted label $F(e(x))$ and select $S_i$ from each group $D_i$. Each $S_i$ contains top $|S_i|$ triplets $x$ based on $F(e(x))$ from $D_i$, i.e.,
\[ |S_i| = \frac{\alpha \cdot |D_i|}{|D|} , \quad
  S_i = \argmax_{V \subset D_i , |V| = |S_i|}  \sum_{x \in V} F(e(x))  .\]
  At last, we combine these $K$ subgroups:
\[  S = S_1 \cup S_2 \cup \ldots \cup S_K , \]

where $S$ is the final high-quality dataset selected by the data selector. The whole selection procedure in the testing stage is shown in Appendix~\ref{detail}.

%% file: sections/experiments.tex
\section{Experimental Setup}

\subsection{Implementation Details}
Our data selector training and subsequent data selection are both on the cc\_sbu\_align dataset~\cite{zhu2023minigpt}, which is used for the second stage fine-tuning in MiniGPT-4 and contains 3439 triplets comprising instructions, images, and responses. 

For the training process, we apply the $K$-means++~\cite{arthur2007k} to split the vision-language dataset into 30 subsets, each containing 114 data points, for acquiring genuine quality labels (detailed in Table~\ref{tab:qualitylab}).
This subset count balances sufficient samples for data selector training and good alignment results for the multimodal model on each subset.
Furthermore, $K$-means++ is employed to ensure that indicators within each subset are similar and differ between subsets. This division strategy guarantees label differentiation for each subset when adjusting the indicators, aiding in data quality assessment.
The data selector is implemented using a self-attention architecture, comprising 2 layers with residual connections. The size of multimodal features concated in the embedding is set to 6. 
The size of the final subset $S$ selected by the data selector is set to $\alpha = 200$, which contains 6\% of the original vision-language instruction data.
Each fine-tuned model is evaluated on the evaluation dataset mentioned in Section~\ref{subsectioneval}. More experimental setting details can be found in Appendix~\ref{subsec:settingdetail}.

\subsection{Evaluation } \label{subsectioneval}
MLLMs are capable of capturing a wide range of multimodal patterns and relationships.
Most are evaluated on publicly available datasets or judged by GPT-4~\cite{openai2023gpt4}. 
Following this trend, we select several popular benchmarks as follows.

We first choose GQA~\cite{hudson2019gqa}, IconQA~\cite{lu2021iconqa}, ScienceQA~\cite{lu2022learn} and OKVQA~\cite{okvqa} to evaluate the MLLMs tuned from different subsets and treat each metrics as the genuine quality labels mentioned in Section~\ref{subsec:qualityeval}. 

Additionally, we test the zero-shot ability of MLLMs on various VQA datasets, including DocVQA~\cite{mathew2021docvqa}, TextVQA~\cite{singh2019towards}, STVQA~\cite{biten2019icdar} and VizWiz~\cite{bigham2010vizwiz}.
We also evaluate the vision and language capabilities in complex multimodal tasks of different models on the recently developed benchmarks including MMBench~\cite{liu2023mmbench} and MME~\cite{fu2023mme}.
Furthermore, we choose GPT-4~\cite{openai2023gpt4} as a judge to compare the responses from MiniGPT-4 and MM-LIMA given the images and instructions from LLaVA-Bench~\cite{liu2023llava}.
The score from GPT-4 is measured by comparing two MLLMs' outputs against a reference answer. 
Detailed description of these evaluation benchmarks are shown in Section~\ref{sec:benchmark}.

\section{Experimental Results}
\subsection{Benchmark Scores} \label{subsection1}
In this section, we conduct quantitative evaluations of MM-LIMA on several datasets using a zero-shot approach. The comparisons of MM-LIMA with the model tuned from 200 random selected samples and MiniGPT-4 are presented in Table~\ref{tab:combined1}, Table~\ref{tab:experiments}, and Table~\ref{tab:combined2}. This assessment offers a valuable perspective on the efficacy of the data selector in enhancing zero-shot performance across a range of tasks.

\begin{figure}[t]
    \centering
    \begin{subfigure}{0.48\textwidth}
        \centering
        \includegraphics[height=4cm]{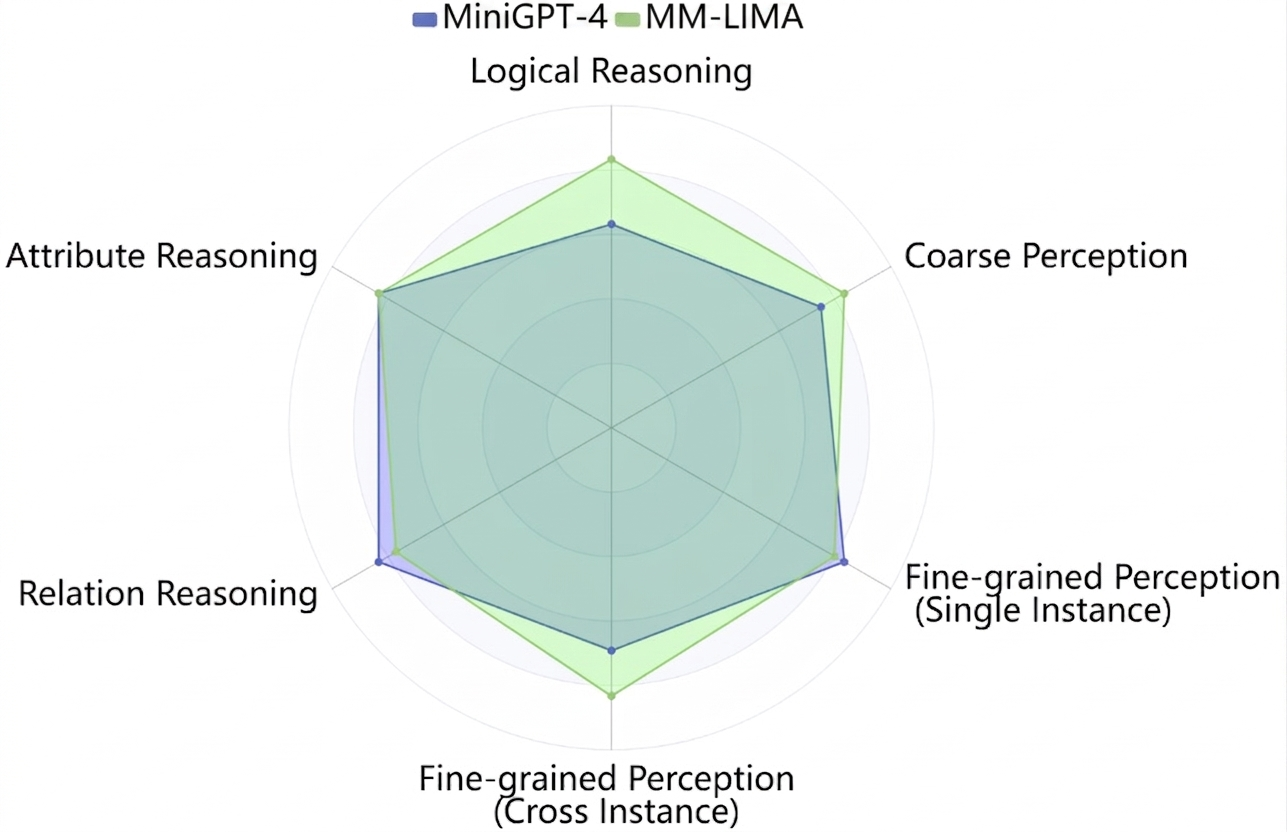}
        \caption{Comparison of 6 dimension levels.}\label{fig:mb1}
    \end{subfigure}
    \begin{subfigure}[b]{0.48\textwidth}
        \centering
        \includegraphics[height=4cm]{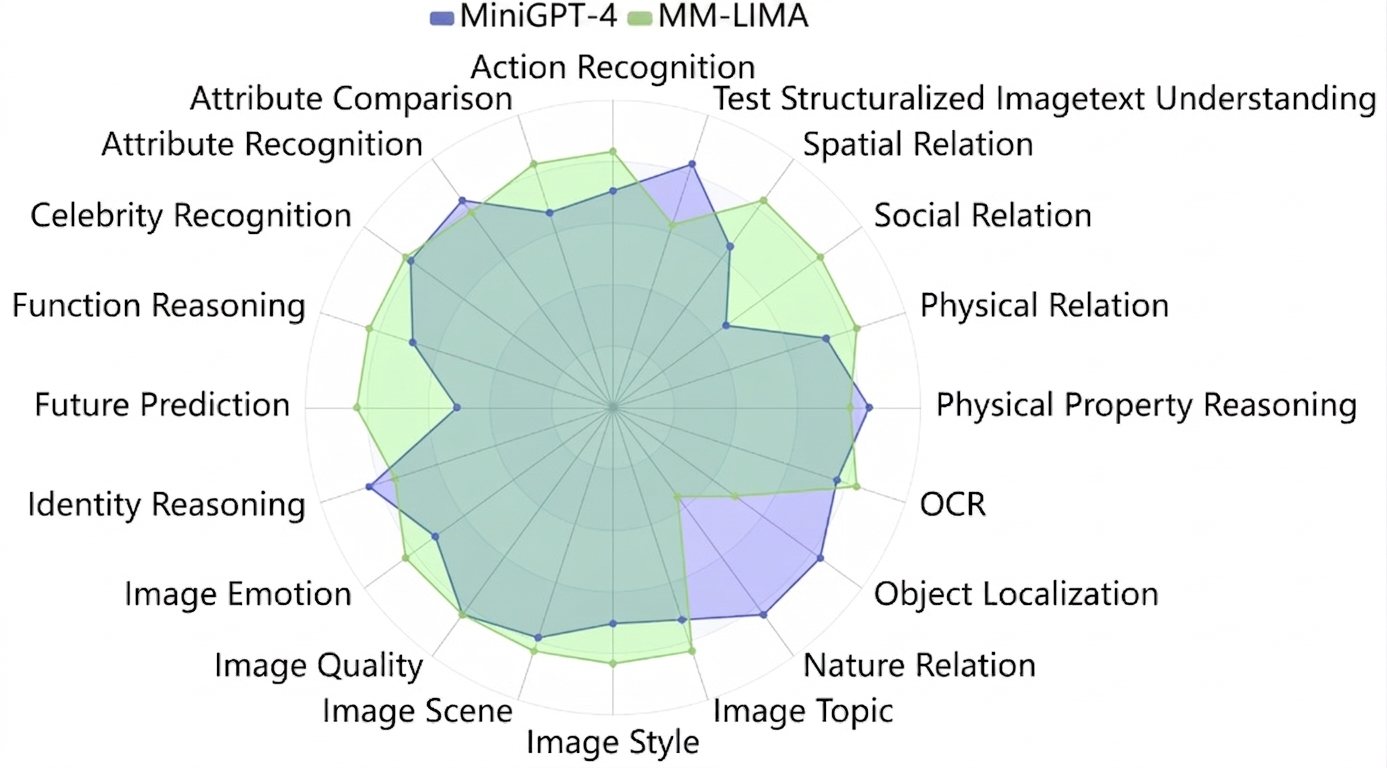}
        \caption{Comparison of 20 ability dimensions.}\label{fig:mb2}
    \end{subfigure}
    \caption{Comparison of MMBench evaluation (MM-LIMA
vs.\ MiniGPT-4). }\label{fig:mb3}
\end{figure}

We observe that MM-LIMA provides the leading performance on average scores in MME (Table~\ref{tab:combined1}) and MMBench (Figure~\ref{fig:mb3} and Table~\ref{tab:combined2}), and realizes transcendence in all aspects of VQA datasets (Table~\ref{tab:experiments}).
Specifically, MM-LIMA demonstrates a +23 score improvement over MiniGPT-4 on MME, +1.55 score on MMBench, and +1.76\% on VQA datasets.
In addition, MM-LIMA outperforms MiniGPT-4 in 8 out of 14 tasks in MME, 4 out of 6 levels as well as 13 out of 20 abilities in MMBench (detailed in Appendix~\ref{mm}), and all 4 VQA datasets in LVLM-eHub.
Moreover, MM-LIMA exceeds the model trained from random samples on all other~tasks.

By evaluating and contrasting these models in a range of tasks, we aim to ascertain the efficacy of our proposed data selector that can effectively identify high-quality data.
Though our data selector is trained on a list of VQA validation sets, MM-LIMA still demonstrates a strong generation ability to out-domain evaluation datasets such as MME and MMBench.
This comprehensive analysis sheds light on the benefits of informed multimodal data selection in enhancing zero-shot performance across diverse and complex tasks.

\begin{table}[t]
\centering
\renewcommand{\arraystretch}{1.1}
    \begin{minipage}[b]{0.53\linewidth}\
    \centering
    \resizebox{\linewidth}{!}{%
    \begin{tabular}{c|c|c|c|c}
    \toprule
    \multicolumn{2}{c|}{\multirow{2}{*}{Evaluation Tasks}}&
     MiniGPT-4
     & Random Selection   & MM-LIMA \\ 
    \multicolumn{2}{c|}{}& (3439 samples)&(200 samples) & (200 samples)\\ 
    \midrule[\heavyrulewidth]
    \multirow{11}*{\rotatebox{90}{Perception}} &
    {Existence} & \textbf{75.00}  & 56.67 $\pm$ 10.00  & 73.33    \\
    
    &{Count} & 30.00  & 23.89 $\pm$ 0.96  & \textbf{31.67}  \\
    
    &{Position} & 36.67 & 33.33 $\pm$ 3.33  & \textbf{38.33}\\
    
    &{Color} & \textbf{38.33} & 30.00 $\pm$ 13.64  & 36.67\\
    
    &{Poster} & 35.71 & 32.99 $\pm$ 1.56 & \textbf{36.05}\\
    
    &{Celerity} & \textbf{62.06} & 50.69 $\pm$ 1.79  & 59.71\\
    
    &{Scene} & 52.75 & 46.00 $\pm$ 2.78 & \textbf{57.75}\\
    
    &{Landmark} & 22.25 & 18.92 $\pm$ 3.55  & \textbf{29.25} \\
    
    &{Artwork} & 23.50 & 16.91 $\pm$ 2.27 & \textbf{50.50} \\
    
    &{OCR} & \textbf{57.50} & 55.00 $\pm$ 4.33 & 50.00\\
    \cmidrule{2-5}
    &{Total Score} & 433.77 & 364.40 $\pm$ 34.92  & \textbf{463.26}\\
    \midrule
    \multirow{5}*{\rotatebox{90}{Cognition}}&{Commonsense Reasoning} & 46.43 & 34.52 $\pm$ 2.70 & 40.00 \\
    
    &{Numerical Calculation} & 47.50 & 45.00 $\pm$ 2.50 & \textbf{50.00} \\
    
    &{Text Translation} & \textbf{50.00} &  44.17 $\pm$ 1.44 & 47.50 \\
    
    &{Code Reasoning} & \textbf{47.50} & 40.17 $\pm$ 3.82 & \textbf{47.50} \\
    \cmidrule{2-5}
    &{Total Score} &  \textbf{191.43} & 162.85 $\pm$ 1.29  & 185.00  \\
    \midrule
    \multicolumn{2}{c|}{Score of MME} &  625.20 & 527.26 %
    & \textbf{648.26} \\
    \bottomrule
    \end{tabular}%
    }
    \captionof{table}{Performance comparison on MME. } 
    \label{tab:combined1}
    \end{minipage}
\hfill
    \begin{minipage}[b]{0.45\linewidth}
    \centering
    \resizebox{\linewidth}{!}{%
    \begin{tabular}{c|c|c|c}
    \toprule
    \multirow{2}{*}{Datasets}&
     MiniGPT-4
     & Random Selection & MM-LIMA \\ 
    & (3439 samples)&(200 samples) & (200 samples)\\ \midrule
    STVQA & 13.71 & 13.51 $\pm$ 0.73 & \textbf{14.55} \\
    VizWiz &  46.60 & 44.47 $\pm$ 0.86 & \textbf{51.02} \\
    DocVQA & 2.77  & 2.58 $\pm$ 0.23 & \textbf{3.01} \\
    TextVQA & 19.06 & 18.92 $\pm$ 0.31 & \textbf{20.62} \\
    \midrule
    Average Score & 20.54 & 19.87 & \textbf{22.30} \\ \bottomrule
    \end{tabular}}
    \captionof{table}{Performance comparison on VQA tasks.}
    \label{tab:experiments}
    \resizebox{\linewidth}{!}{%
    \begin{tabular}{c|c|c|c}
    \toprule
    \multirow{2}{*}{Dimension Level}&
     MiniGPT-4
     & Random Selection  & MM-LIMA \\ 
    & (3439 samples)&(200 samples)  & (200 samples)\\ 
    \midrule[\heavyrulewidth]
    LR & 12.50  & 17.23 $\pm$ 1.76 & \textbf{16.48}    \\
    
    AR & \textbf{41.87}  & 35.41 $\pm$ 5.09 & \textbf{41.87}  \\
    
    RR & \textbf{12.68} & 9.23 $\pm$ 2.11 & 11.74\\
    
    FP-C & 17.60 & 13.20 $\pm$ 3.39 & \textbf{21.20}\\
    
    FP-S & \textbf{35.75} & 26.92 $\pm$ 4.14 & 34.25\\
    
    CP & 38.30 & 29.29 $\pm$ 5.55 &\textbf{42.55} \\
    \midrule
    Score of MMBench&  29.87 & 23.95 & \textbf{31.42}  \\
    \bottomrule
    \end{tabular}%
    }
    \captionof{table}{Performance comparison on MMBench.}
    \label{tab:combined2}
    \end{minipage}
\end{table}

\subsection{GPT-4 Evaluation}

\begin{wrapfigure}{r}[-0.5cm]{0.29\linewidth}
  \centering
  \vspace{-0.5cm}
  \includegraphics[width=0.8\linewidth]{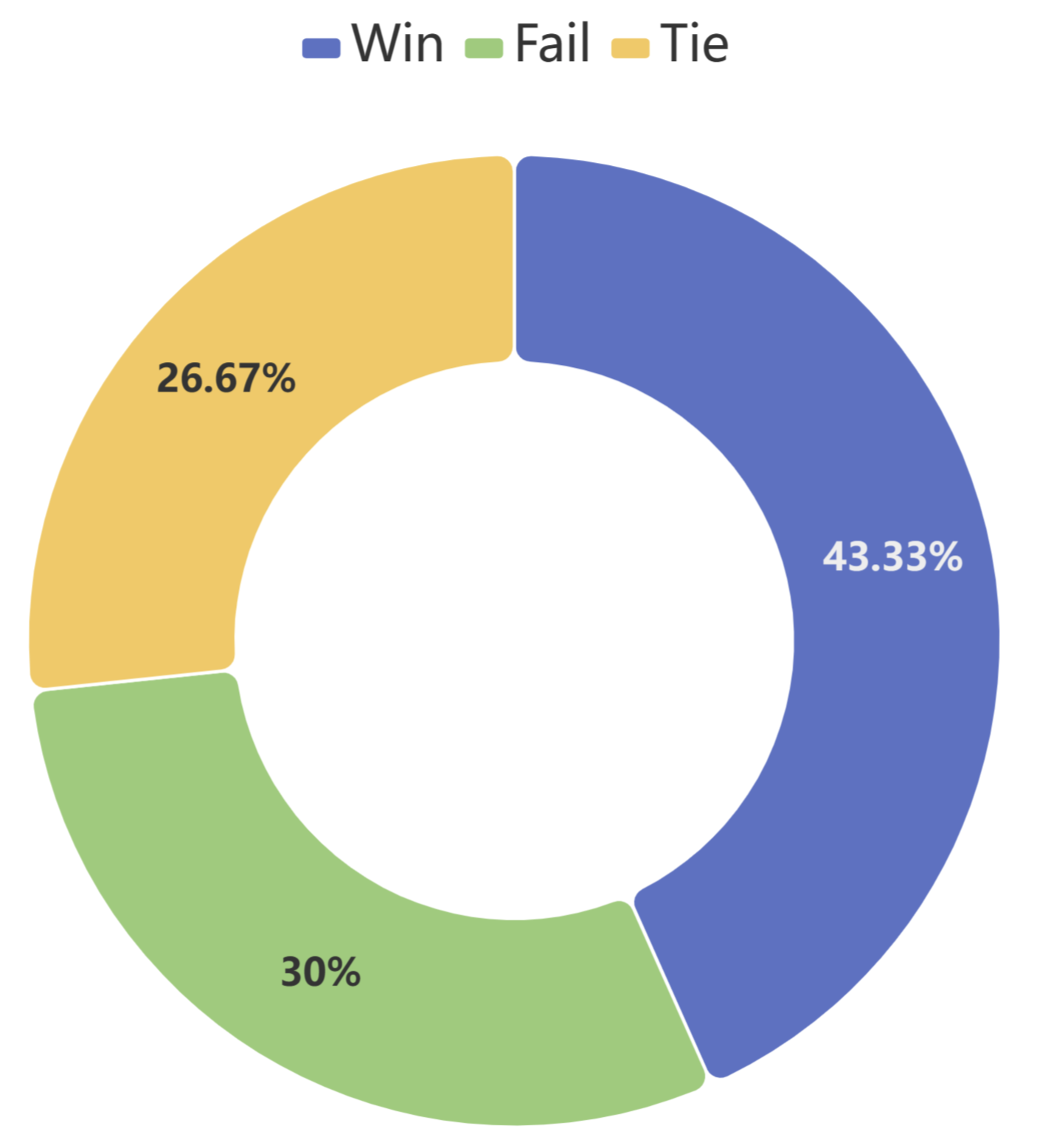} 
  \caption{GPT-4 Evaluation Comparison ({MM-LIMA} vs.\ {MiniGPT-4}).} 
  \label{fig:compare} 
  \vspace{-1cm}
\end{wrapfigure}

Given the presence of inherent position bias within LLMs as evaluators, wherein certain positions are favored over others \cite{wang2023large}, we have undertaken measures to address this concern.
To mitigate such bias, we conduct evaluations using both response orders -- placing MM-LIMA's generated response before and after MiniGPT-4's response.
To establish a definitive judgment criterion, we introduce the ``Win-Tie-Fail'' framework, characterized as follows:

    1) Win: MM-LIMA is deemed the winner in two instances, or secures victory once and achieves a draw once;
    2) Tie: MM-LIMA achieves a draw twice, or prevails in one instance and succumbs in another;
    3) Fail: MM-LIMA faces defeat in two instances, or experiences a loss once and attains a draw once.
    
    The results of this evaluation are depicted in Figure~\ref{fig:compare}. Win, Fail, and Tie in this figure denote comparative outcomes when the generation results of MM-LIMA are evaluated against those of MiniGPT-4.
    Throughout 60 questions, MM-LIMA emerges victories in 26 instances, experiences failure in 16, and achieves a tie in 18.
    This evidence underscores the notable superiority of MM-LIMA's response quality in comparison to MiniGPT-4.

\subsection{Demonstrations}
We conducted a comparative assessment of image understanding and conversation abilities between MM-LIMA and MiniGPT-4, focusing on a challenging instance described in Table~\ref{tab:story} and Table~\ref{tab:cot} in Appendix~\ref{appendixdemo}. Despite being fine-tuned on a smaller multimodal instruction dataset, MM-LIMA outperforms MiniGPT-4 in complex image understanding and reasoning tasks. This highlights MM-LIMA's exceptional ability for advanced reasoning, emphasizing its prowess in image comprehension and executing instruction-following tasks.

\subsection{Ablation Study} \label{subsection2}
Through a series of ablation studies, we elucidate the contributions of various factors, including different indicators, diverse data selector architectures, clustering for diversity, and selected data size, to the overall effectiveness of our data selection approach. 
Our experimental analyses serve as empirical verification of the theoretical foundations we have put forward.

\noindent\textbf{Analysis of Different Indicators.}~~
To comprehensively evaluate the impact of distinct indicators on the data selection process, we conduct another ablation study.
Each individual indicator is isolated and its effect on 200 data selection is scrutinized.
As showcased in the left part of Table~\ref{tab:diin}, the models fine-tuned using data selected based on CLIP Score, Length Score, Reward Score, GPT Score, and Multimodal Features consistently outperform those generated through random sampling.
This illustrates that employing each separate indicator yields positive effects on the data selection process, thus they are suitable for data quality assessment.

\begin{minipage}[b]{1\linewidth}
\centering
\renewcommand{\arraystretch}{1.2}
\resizebox{0.8\textwidth}{!}{%
\begin{tabular}{c|cccccc|ccc}
\toprule
\multirow{2}{*}{Benchmark}& \multicolumn{6}{c|}{Indicators} & \multicolumn{3}{c}{Self-Attention Layers} \\ \cmidrule{2-10}
 & Random & CLIP  & Reward  & Length & GPT & Features& 1 & 2  & 3 \\ 
 \midrule
MME &  $527.26$ & $95.79\uparrow$ & $33.36\uparrow$ & $90.76\uparrow$ & $59.69\uparrow$ & $20.82\uparrow$ &  $521.10$ & $\mathbf{648.26}$ & $594.84$\\
\midrule
VQA &  $19.87$ & $2.57\uparrow$ & $1.47\uparrow$ & $2.39\uparrow$ & $2.11\uparrow$ & $0.63\uparrow$ &  $22.08$ & $\mathbf{22.30}$ & $21.74$\\
\bottomrule
\end{tabular}%
}
\captionof{table}{MME and VQA scores under different indicators used separately (left) and different self-attention layers (right). }%
\label{tab:diin}

  \end{minipage}

\begin{figure}[t]
    \centering
    \begin{subfigure}[b]{0.47\textwidth} 
        \includegraphics[width=\textwidth, trim= 0 50 0 0, clip]{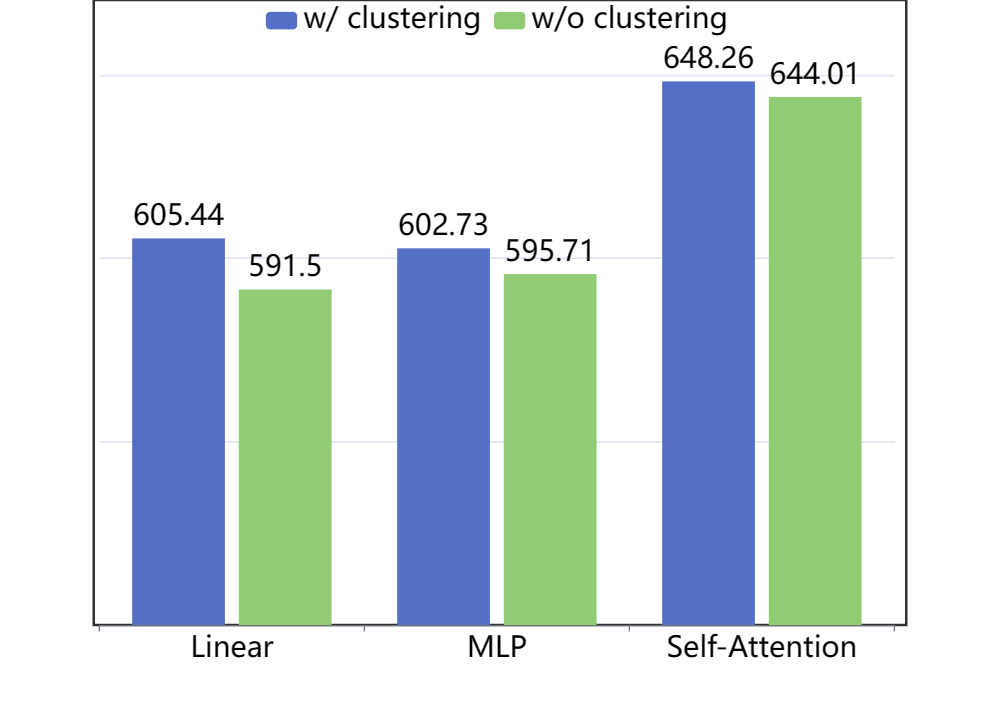}
        \caption{Comparison of MME evaluation.} 
        \label{fig:subfig11} 
    \end{subfigure}
    \begin{subfigure}[b]{0.47\textwidth}
        \includegraphics[width=\textwidth, trim= 0 50 0 0, clip]{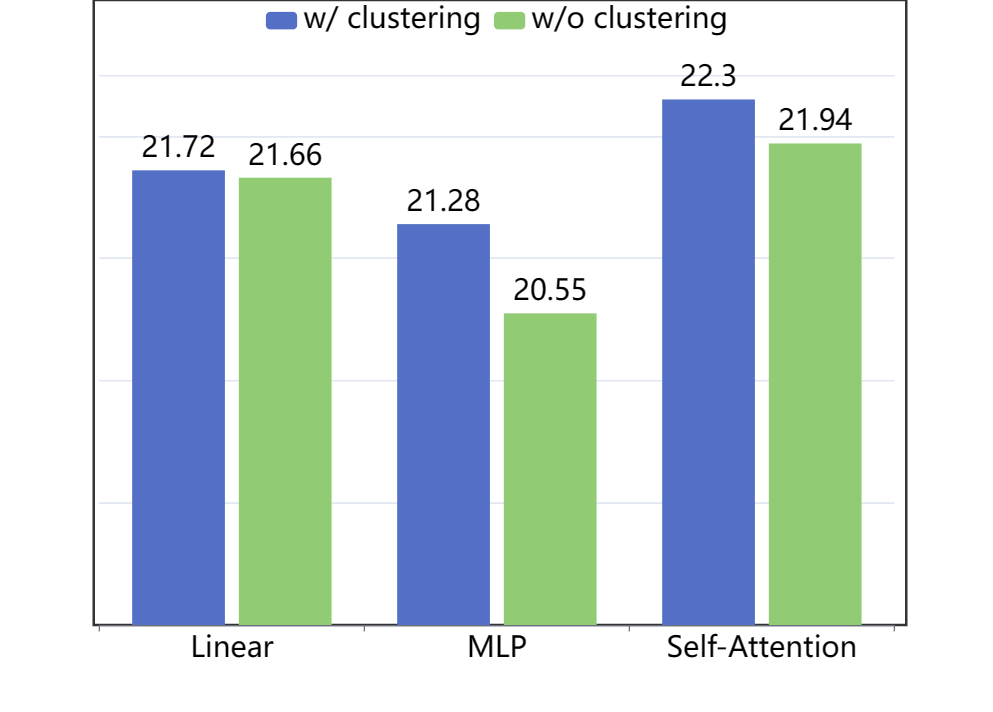} 
        \caption{Comparison of VQA evaluation.} 
        \label{fig:subfig22}
    \end{subfigure} 

    \caption{Ablation study to investigate the impact of clustering in the testing stage and different types of network structures utilized in the data selector. Note that self-attention with clustering consistently yields leading performance.}\vspace{-15pt}
    \label{fig:sel_type}
\end{figure}

\noindent\textbf{Analysis of the Data Selector Architecture.}~~
In this ablation study, we try three different structures for data selector, including linear regression, MLP, and self-attention. Compared to MLP or linear models, self-attention mechanisms enable internal information interaction within embeddings, ensuring mutual awareness between features and scores within the embedding. In contrast, MLP or linear models can only achieve global awareness. The results depicted in Figure~\ref{fig:sel_type} indicate that the MLLM fine-tuned using data selected by the self-attention structure achieves the highest performance. This suggests that the self-attention mechanism is more adept at learning the mapping relationship between embeddings and genuine quality labels, thereby enabling the data selector to better identify high-quality data.
In addition, we conduct experiments with different numbers of attention layers, as summarized in the right part of Table~\ref{tab:diin}. Notably, we find that employing 2 layers is the most suitable configuration, as it resulted in optimal performance.

\noindent\textbf{Analysis of Clustering.}~~
The application of spectral clustering within the data selector mechanism ensures the diversity of the chosen vision-language instruction data.
To dissect the contribution of clustering, we conduct an ablation study by removing the clustering mechanism.
The results are presented in Figure~\ref{fig:sel_type} by comparing the two bars for each data selector structure. Incorporating clustering into data selectors with different structures consistently yields improved performance, highlighting the significance of clustering in enhancing the fine-tuning procedure. 

\begin{figure}[t]
    \centering
    \includegraphics[width=0.35\textwidth, trim=50 0 50 0,clip]{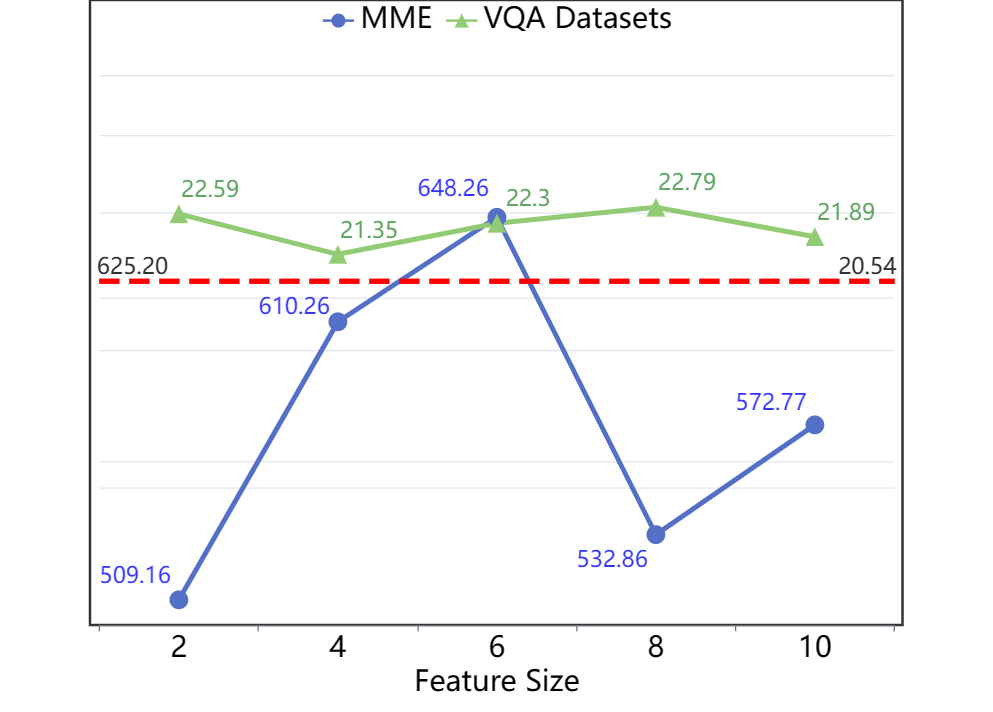}
    \includegraphics[width=0.35\textwidth, trim=50 0 50 0,clip]{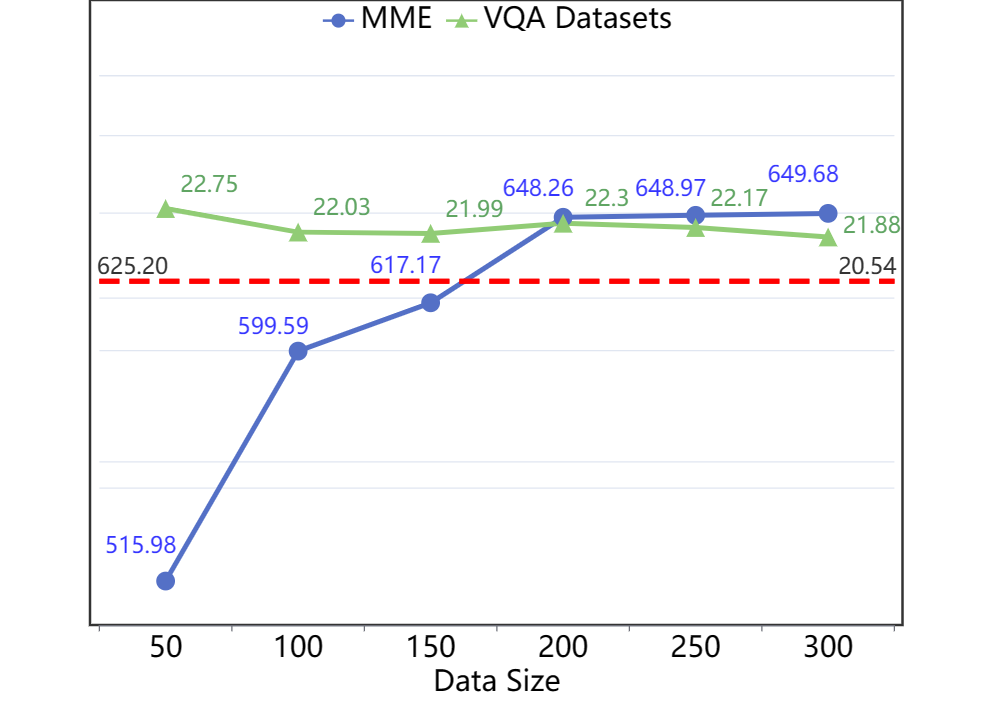}
  \caption{The left part denotes different multimodal feature sizes. The right part denotes different curated data sizes. The red dotted lines represent the performance of MiniGPT-4.}
    \label{fig:sizes}
\end{figure}

\noindent\textbf{Analysis of Multimodal Feature Size.}~~
We also explore various sizes of multimodal features after conducting unsupervised dimensionality reduction. The results presented on the left side of Figure~\ref{fig:sizes} demonstrate that setting this size to 6 consistently yields the best performance.
Our analysis indicates that when dimensionality reduction is configured with a low value, it may excessively compress multimodal features. Conversely, if dimensionality reduction is set too high, it can lead to an expansion in the embedding's dimensionality, thereby increasing the number of parameters that the data selector needs to learn. This heightened dimensionality can make the data selector more susceptible to overfitting or underfitting issues.

\noindent\textbf{Analysis of Selected Data Size.}~~
We aim to identify the minimum amount of data required to make MM-LIMA surpass MiniGPT-4. From the right side of Figure~\ref{fig:sizes}, we discover that selecting 50 data points is sufficient when considering the VQA dataset for evaluation. However, due to the gap between the datasets used for evaluation, such as MME and MMBench, 200 data points are needed for fine-tuning the model to comprehensively outperform MiniGPT-4.  This observation underscores the strong transferability of our designed data selector.

%% file: sections/discussion.tex
\section{Related Works}

\textbf{Visual Instruction Tuning.}
Instruction tuning is a learning paradigm that fine-tunes pre-trained LLMs on datasets described by natural language instructions.
Through this training method, the zero-shot abilities of LLMs can be significantly enhanced.
The effectiveness of instruction tuning has been demonstrated by a set of research, including FLAN~\cite{wei2021finetuned}, InstructGPT~\cite{ouyang2022training}, and ChatGPT.
Inspired by this, several recent works aim at enabling LLMs to handle multimodal tasks with visual instruction tuning, such as MiniGPT-4~\cite{zhu2023minigpt}, LLaVA~\cite{liu2023llava},  LLaMA-Adapter~\cite{zhang2023llama} and InstructBLIP~\cite{dai2023instructblip}.
These works choose linear projection layers as the bridges between image encoders and LLMs, and perform visual instruction tuning either on self-instruct datasets~\cite{zhu2023minigpt,liu2023llava} or on existing multimodal datasets~\cite{zhang2023llama,dai2023instructblip}.

\textbf{Instruction Curation.}
To improve model performance after instruction tuning, some relevant works manage to filter low-quality instruction data or construct carefully curated examples during the fine-tuning stage, thereby enhancing model capabilities.
LIMA~\cite{zhou2023lima} shows that fine-tuning a strong pre-trained language model on 1000 human curated and high-quality examples can produce remarkable, competitive results on a wide range of prompts. 
Several recent works~\cite{paul2021deep, cao2023instruction, chen2023alpagasus} have developed instruction quality evaluation methods for measuring the quality of vision or language datasets, such as computing the perplexity, calculating the gradient, and acquiring ChatGPT for rating, to filter low-quality data for training.
In particular, Instruction Mining~\cite{cao2023instruction} proposes a linear rule for selecting high-quality instruction data, which does not require human annotation in LIMA.
Unlike the aforementioned works in a single modal, MM-LIMA is the first multimodal large language model that achieves better performance through effective data selection. In contrast to Instruction Mining~\cite{cao2023instruction}, our work aims to present a more general data quality evaluation pipeline and train a robust data selector to automatically select proper data from the raw dataset used during fine-tuning.

\section{Conclusion}
In this paper, we provide a thorough analysis of the data selector's effectiveness in curating valuable multimodal instruction data. We also extensively evaluate MM-LIMA's performance on various datasets, confirming its excellence in producing coherent and accurate outputs.
Our research's core thesis emphasizes the power of carefully chosen less but high-quality instruction data in enhancing multimodal large language models' performance. MM-LIMA's success underscores that inducing instruction data by proper selection can lead to significant advancements in the field of multimodal LLMs, fostering improved instruction understanding and generation capabilities.

%% file: sections/appendix.tex
\begin{center}
    \Large \textbf{Appendix}\\[1cm]
\end{center}

\section{Implementation Details of MM-LIMA} 
In this section, we first present the design of our GPT-4 prompt for rating scores and the genuine quality labels for training the data selector. We then provide an implementation of our data selection algorithm and further experimental settings.

\subsection{GPT Prompt} \label{gptprompt}
We provide the detailed prompt to GPT-4 used for rating scores in Table~\ref{tab:prompt}. It is similar to the prompt for rating and filtering training data in Alpagasus~\cite{chen2023alpagasus}.

\begin{table}[H]
\centering
\resizebox{\linewidth}{!}{
\begin{tabular}{cm{14cm}}
\toprule[2pt]
\multicolumn{2}{c}{GPT Prompt}\\         
\midrule[2pt]
{System Prompt} & We would like to request your feedback on the performance of an AI assistant. The assistant provides a caption based on an image and an instruction. 

Instruction: [Instruction]

Caption: [Caption] \\ 
\midrule
{User Prompt} & Please rate according to the quality and variety of the caption to the instruction. Each assistant receives a score on a scale of 0 to 100, where a higher score indicates higher level of the quality and variety. Please first output a single line containing the value indicating the scores. In the subsequent line, please provide a comprehensive explanation of your evaluation, avoiding any potential bias. The instruction and caption are displayed following without image. \\ \bottomrule[2pt]
\end{tabular}}
\caption{Prompt $p_G$ to GPT-4 for rating multimodal data.} \label{tab:prompt}
\end{table}

\subsection{Genuine Quality Labels} \label{gelabel}
To acquire genuine quality labels for the data, we choose to partition the cc\_sbu\_align dataset into 30 subsets using clustering techniques such as $K$-means++~\cite{arthur2007k}. Each of these subsets, denoted as $i \in \{1, 2, \dots, 30\}$, comprises 114 data points. Subsequently, each subset is employed to fine-tune a pre-trained Multimodal Language Model (MLLM). These fine-tuned models are then evaluated on a validation set, including GQA~\cite{hudson2019gqa}, IconQA~\cite{lu2021iconqa}, OKVQA~\cite{okvqa} and ScienceQA~\cite{lu2022learn}, to generate scores in Table~\ref{tab:qualitylab}, which serves as genuine quality labels for the respective subset.

\begin{table}[t]
\centering
\renewcommand{\arraystretch}{1.1}

\resizebox{0.52\textwidth}{!}{%
\begin{tabular}{c|c|c|c|c|c}
\toprule
{Subset}& GQA & IconQA  & OKVQA & ScienceQA & Average\\ 

\midrule[\heavyrulewidth]
1 & 28.48  &35.88 & 37.11 & 21.98  & 30.86\\

2 & 29.17  & 37.78 & 35.85 & 21.29 & 31.02\\

3 & 27.21 & 35.35 & 33.83 & 21.63 & 29.51\\

4 & 28.13  &35.64  &36.77  &21.75  &30.57\\

5 & 28.25  & 35.75 &36.26  &23.56  &30.95\\

6 & 28.72  & 35.92 & 35.67 &22.46  &30.69\\

7 & 28.08  &35.18  &36.24  &22.28  &30.45\\

8 & 28.20  &35.71  &36.21  &22.06  &30.55\\

9 & 28.49  &36.79  &37.42  &23.32  &31.51\\

10 & 27.40  &37.89  &35.71  &23.78  &31.20\\

11 & 27.84  &37.57  &36.39  &23.65  &31.36\\

12 & 30.68  & 35.36 & 36.49 & 21.60 &31.03\\

13 & 27.68  &38.64  &36.78  &23.65  &31.69\\

14 &  29.00 & 37.23 & 36.69 &23.85  &31.69\\

15 & 28.31  &38.03  &36.00  &23.90  &31.56\\

16 & 28.82  &34.91  &35.24  &21.40  &30.09\\

17 & 27.10  &35.14  &35.02  &22.44  &29.93\\

18 &  27.70 &35.76  &36.02  &22.03  &30.38\\

19 &  29.33 & 37.04 & 35.98 &23.19  &31.38\\

20 &28.75   &36.58  &35.92  &22.59  &30.96\\

21 &  29.67 & 36.33 & 36.52 &22.91  &31.36\\

22 & 27.68  &36.67  &36.27  &22.05  &30.67\\

23 & 28.68  &36.54  & 36.47 &23.63  &31.33\\

24 & 29.45  &37.31  &35.98  &22.74  &31.37\\

25 & 26.77  &35.50  &34.69  &22.71  &29.92\\

26 & 28.62  &34.90  &35.00  &22.08  &30.15\\

27 & 26.52  &35.71  &34.21  &23.00  &29.86\\

28 & 27.48  &36.53  &35.58  &22.97  &30.64\\

29 & 27.93  &34.54  &36.00  &22.79  &30.31\\

30 & 28.53  &36.98  &37.34  &23.68  &31.64\\

\bottomrule
\end{tabular}%
}
\caption{30 genuine quality labels.}
\label{tab:qualitylab}
\end{table}

\subsection{Selection Algorithm Implementation Details} \label{detail}

After finishing training the data selector, we proceed to split the multimodal dataset for fine-tuning into $K = 10$ groups for data selection.
This division is achieved by employing spectral clustering on the image embeddings, which have been encoded using ViT from CLIP~\cite{radford2021learning}.
The purpose of this clustering step is to ensure diversity within our selected data, as it helps capture a wide range of data distribution patterns. 
It's important to note that the clustering algorithm used here (denoted as $\Lambda$) is distinct from the one used earlier (denoted as $\Gamma$) for dividing the subsets and obtaining genuine quality labels. Each subset created using $\Gamma$ shares the same genuine quality labels. 
By introducing this differentiation between $\Lambda$ and $\Gamma$ for data selection after training the data selector, we ensure that predicted labels for data points within each cluster maintain their distinctiveness, preventing potential label confusion.
\begin{algorithm}[t]
\caption{\textsc{data selection}}
\label{alg:dataset_selection}
\begin{algorithmic}[1]
\REQUIRE Dataset $D$, Trained Data Selector $F$, number of clusters $K$, subset size factor $\alpha$
\STATE Compute clusters $D_1, D_2, \ldots, D_K$ using a clustering algorithm $\lambda$ on images in $D$

\FOR{$i=1$ to $K$}
    \FOR{$x$ in $D_i$}
        \STATE Compute CLIP Score $C(x)$, Length Score $L(x)$, Reward Score $R(x)$, GPT Score $G(x, p_G)$ and Multimodal Features $P(f(x_{\text{image}}), g(x_{\text{response}}))$
        \STATE Concat the indicators as embedding $e(x)$ in Equation~\eqref{eq:embed}
        \STATE Compute predicted label $F(e(x))$
    \ENDFOR
    \STATE Compute $|S_i| = \frac{\alpha \cdot |D_i|}{|D|}$
    \STATE Select top $|S_i|$ samples from $D_i$ based on $F(e(x))$ to form $S_i$
\ENDFOR

\STATE Combine $S_1, S_2, \ldots, S_K$ to form $S$

\RETURN $S$
\end{algorithmic}
\end{algorithm}

\subsection{Experimental Setting Details} \label{subsec:settingdetail}
When computing the indicators for each triplet containing images, instructions, and responses, we do not take instructions into consideration because they have fixed formats in the dataset (e.g., ``Describe this image in detail.'').

To generate multimodal features, we employ Principal Component Analysis (PCA) to reduce the dimensionality of the multimodal features generated by frozen ViT~\cite{ilharco_gabriel_2021_5143773} and Llama2~\cite{touvron2023llama}.

For the data selector training, we set the number of training epochs to 20 and the learning rate to 0.01.
We conduct all instruction tuning on pre-trained 7B MiniGPT-4 \cite{zhu2023minigpt} and use the same fine-tuning hyperparameters as the original MiniGPT-4.

For the ablation study of different indicators, we follow the testing stage of the data selector and sort each data point after conducting clustering. We sort each data point based on CLIP Score, Reward Score, and GPT Score respectively. Additionally, we select a length range to evaluate the length effect. Besides, we train another self-attention network with multimodal features as inputs for data selection.

\subsection{Evaluation Benchmarks}\label{sec:benchmark}
\textbf{MME}~\cite{fu2023mme}.
It is an MLLM evaluation benchmark that measures both perception and cognition abilities on a total of 14 subtasks. The full score for the overall tasks is 2800, while for the subtasks is 200.
For each test image, MME adopts an instruction of a question and a description “Please answer yes or no” to prompt MLLMs. Such a concise instruction-answer evaluation allows for a fair comparison of MLLMs without the impact of prompt engineering.

\textbf{MMBench}~\cite{liu2023mmbench}.
This benchmark is collected from multiple sources, including public datasets and Internet, and currently, contains 2974 multiple-choice questions, covering 20 ability dimensions. The existing 20 ability dimensions are structured into 6 dimension levels.  
Each question is a multiple-choice format with a single correct answer. 
For a more reliable evaluation, it employs ChatGPT to match a model's prediction with the choices of a question, and then output the corresponding label (A, B, C, D) as the final prediction.

\textbf{VQA Datasets}.
LVLM-eHub~\cite{xu2023lvlmehub} is a comprehensive evaluation benchmark for publicly available MLLMs.
Based on this platform, we choose GQA~\cite{hudson2019gqa}, IconQA~\cite{lu2021iconqa}, ScienceQA~\cite{lu2022learn} and OKVQA~\cite{okvqa} to evaluate the MLLMs tuned from different subsets and treat each metrics as the genuine quality labels mentioned in Section~\ref{subsec:qualityeval}. We also test the zero-shot ability of MLLMs on various datasets, including DocVQA~\cite{mathew2021docvqa}, TextVQA~\cite{singh2019towards}, STVQA~\cite{biten2019icdar} and VizWiz~\cite{bigham2010vizwiz}.
Top-1 accuracy is employed for these tasks.

\textbf{LLaVA-Bench}~\cite{liu2023llava}.
It collects a diverse set of 24 images with 60 questions in total, including indoor and outdoor scenes, memes, paintings, sketches, etc.
It associates each image with a highly detailed and manually curated description and a proper selection of questions that are categorized into conversation (simple QA), detailed description, and complex reasoning.
We choose GPT-4 as a judge to compare the responses from MiniGPT-4 and MM-LIMA given the images and instructions from LLaVA-Bench.
The score from GPT-4 is measured by comparing two MLLMs' outputs against a reference answer. 
Such a design assesses the model's robustness to different prompts. 

\section{More Experimental Results} \label{results}
\subsection{MMBench Results} \label{mm}
MMBench~\cite{liu2023mmbench} gathers approximately 3000 questions spanning 20 ability dimensions in 6 levels. 
The detailed assessment of 20 abilities is illustrated in Table~\ref{tab:mb3}.

\begin{table}[h]
    \centering
\renewcommand{\arraystretch}{1.1}

\resizebox{0.9\textwidth}{!}{%
\begin{tabular}{c|c|c|cc}
\toprule
\multirow{2}{*}{Ability Dimension}&
 MiniGPT-4
 & Random Selection  & MM-LIMA \\ 
& (3439 samples)&(200 samples)  & (200 samples)\\ 
\midrule[\heavyrulewidth]
Action Recognition & 37.50  &28.03 $\pm$ 6.84 & \textbf{44.32}    \\

Attribute Comparison & 5.00  & 3.75 $\pm$ 2.04 & \textbf{6.25}  \\

Attribute Recognition & \textbf{51.00} & 41.33 $\pm$ 2.87 & 48.00\\

Celebrity Recognition & 34.75 & 22.03 $\pm$ 7.22 & \textbf{35.59}\\

Function Reasoning & 34.58 & 29.91 $\pm$ 6.05 & \textbf{42.06}\\

Future Prediction & 18.92 & \textbf{34.23} $\pm$ 2.78 &31.08 \\

Identity Reasoning & \textbf{69.51} & 56.91 $\pm$ 6.40 & 62.20 \\

Image Emotion & 35.71 & 38.89 $\pm$ 1.48 &\textbf{41.67} \\

Image Quality & \textbf{3.49} & 1.94 $\pm$ 1.45 &\textbf{3.49} \\

Image Scene  & 66.15 & 49.74 $\pm$ 11.17 &\textbf{70.00} \\

Image Style & 31.76 & 19.22 $\pm$ 9.72 &\textbf{37.65} \\

Image Topic  & 40.00 & 26.27 $\pm$ 4.93 &\textbf{45.88} \\

 Nature Relation & \textbf{17.28} & 6.17 $\pm$ 1.01 &7.41 \\
 
Object Localization & \textbf{16.19}& 12.38 $\pm$ 0.78 &9.52 \\

OCR  & 44.16 & 35.50 $\pm$ 6.12 &\textbf{48.05} \\

 Physical Property Reasoning & \textbf{27.00} & 23.67 $\pm$ 3.09 &25.00 \\
 
Physical Relation    & 13.46 & 14.10 $\pm$ 3.27 &\textbf{15.38} \\

 Social Relation& 7.50 & 9.17 $\pm$ 4.13 &\textbf{13.75} \\
 
Spatial Relation & 8.54 & 6.50 $\pm$ 1.52 &\textbf{10.98} \\

 Test Structuralized Imagetext Understanding   & \textbf{7.84} & 4.90 $\pm$ 1.60 &5.88\\
\midrule
Score of MMBench&  29.87 & 23.95 & \textbf{31.42}  \\
\bottomrule
\end{tabular}%
}
\captionof{table}{Performance comparison on MMBench.}
\label{tab:mb3}
\end{table}

\subsection{Additional Demonstrations} \label{appendixdemo}
We display both MM-LIMA and MiniGPT-4 multimodal chatbot demos below.

Table \ref{tab:story} and Table \ref{tab:cot} distinctly indicate that MM-LIMA possesses the capacity to generate responses that are not only more comprehensive but also exhibit a higher level of fluency when compared to those produced by MiniGPT-4.

The story presented by MM-LIMA in Table~\ref{tab:story} is vivid and detailed, which involves the key information from the image. Furthermore, when tasked with writing a recipe based solely on a food image in Table \ref{tab:cot}, MM-LIMA demonstrates a notably superior ability in chain-of-thought by generating robust and effective responses step by step.
These findings collectively emphasize the enhanced language generation prowess of MM-LIMA, which contributes to more detailed, expressive, and contextually appropriate responses in various scenarios.

\begin{table}[t]
\centering
\resizebox{\linewidth}{!}{
\begin{tabular}{cm{18cm}}
\toprule[2pt]
\multicolumn{2}{c}{Demos}         \\ 

\multicolumn{2}{c}{\makecell{\includegraphics[width=0.35\textwidth]{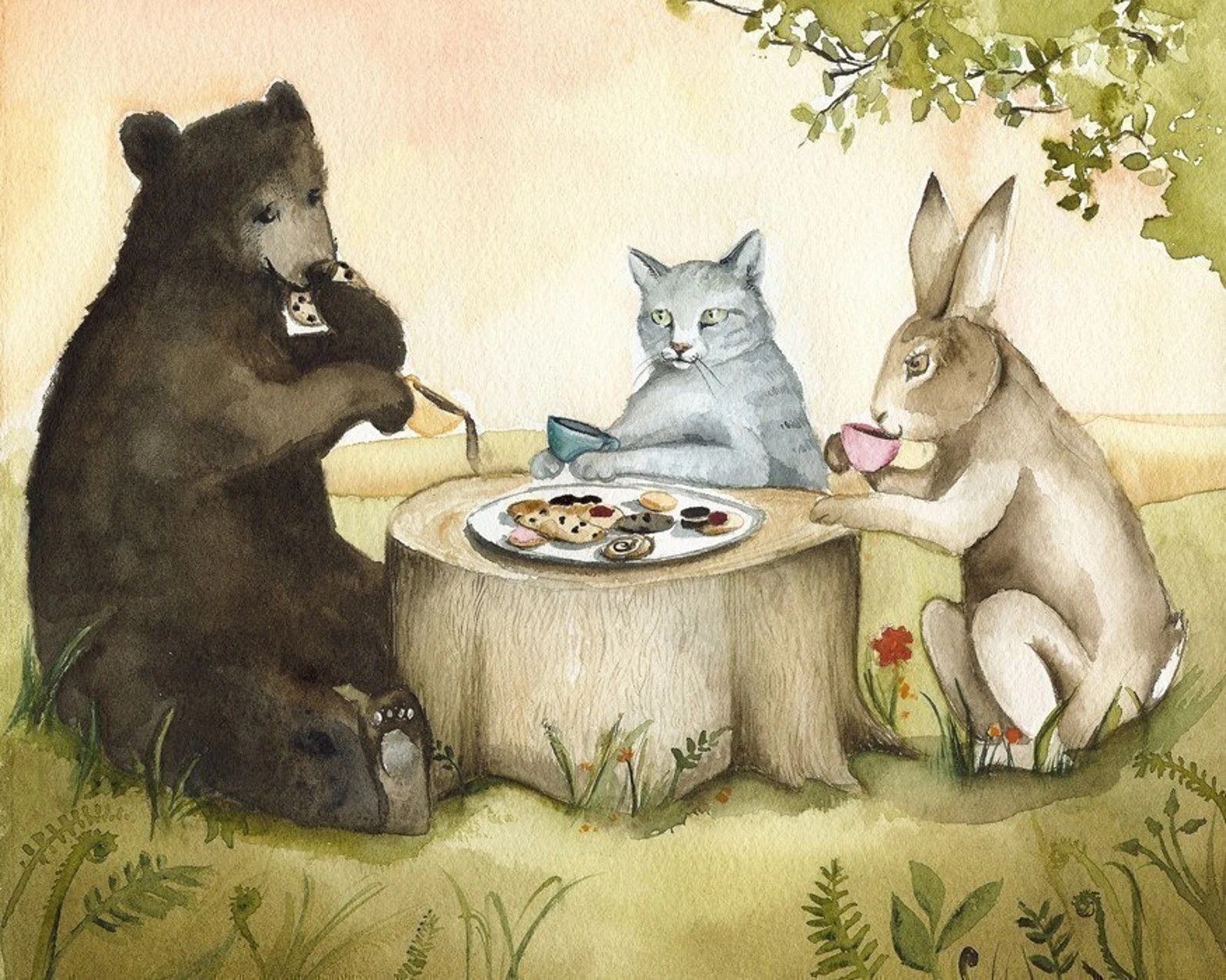}}}                  \\ 

\midrule[2pt]
\multicolumn{1}{c}{User}      & Can you write a story around this image for my child? \\ \midrule[2pt]
\multicolumn{1}{c}{MM-LIMA} &
Yes, I can write a story around this image.

Once upon a time, there was a group of animals who lived in a forest together. The animals were a bear, a rabbit, and a cat. They all loved to sit down to a meal together, especially for breakfast.

One morning, the bear woke up early and decided to make breakfast for his friends. He went to the stream and caught some fish, then returned to the table to cook them in a pan with butter. The rabbit helped him by bringing him some eggs, and the cat helped by bringing him some cream.

They all sat down at the table under the shade of a tree and enjoyed their breakfast together. The bear said, “This is the best breakfast I’ve ever had!” The rabbit said, “I’m glad you like it, Bear. This is my favorite too.” The cat said, “Meow, meow! This is delicious.”

The animals enjoyed their meal and their friendship. They promised to have breakfast together every morning for as long as they lived in the forest. And so they did, until the day they all had to say goodbye to each other, as they each had to go their own way.

But the memories of their breakfasts together remained, and they were always happy to think back on the good times they had in the forest.
\\ \midrule
\multicolumn{1}{c}{MiniGPT-4} & Once upon a time, there was a bear, a rabbit, and a cat who lived in the forest. They spent their days exploring the forest, playing games, and having picnics together. One day, they decided to have a picnic at a table in the middle of the forest. They sat around the table, enjoying their meal and the beautiful scenery around them. The bear ate his share of the food, the rabbit ate her share, and the cat drank some milk. They were all happy and content. \\ \bottomrule[2pt]
\end{tabular}}
\caption{An Story Generation Example.} \label{tab:story}
\end{table}

\begin{table}[t]
\centering
\resizebox{\linewidth}{!}{
\begin{tabular}{cm{18cm}}
\toprule[2pt]
\multicolumn{2}{c}{Demos}         \\ 

\multicolumn{2}{c}{\makecell{\includegraphics[width=0.35\textwidth]{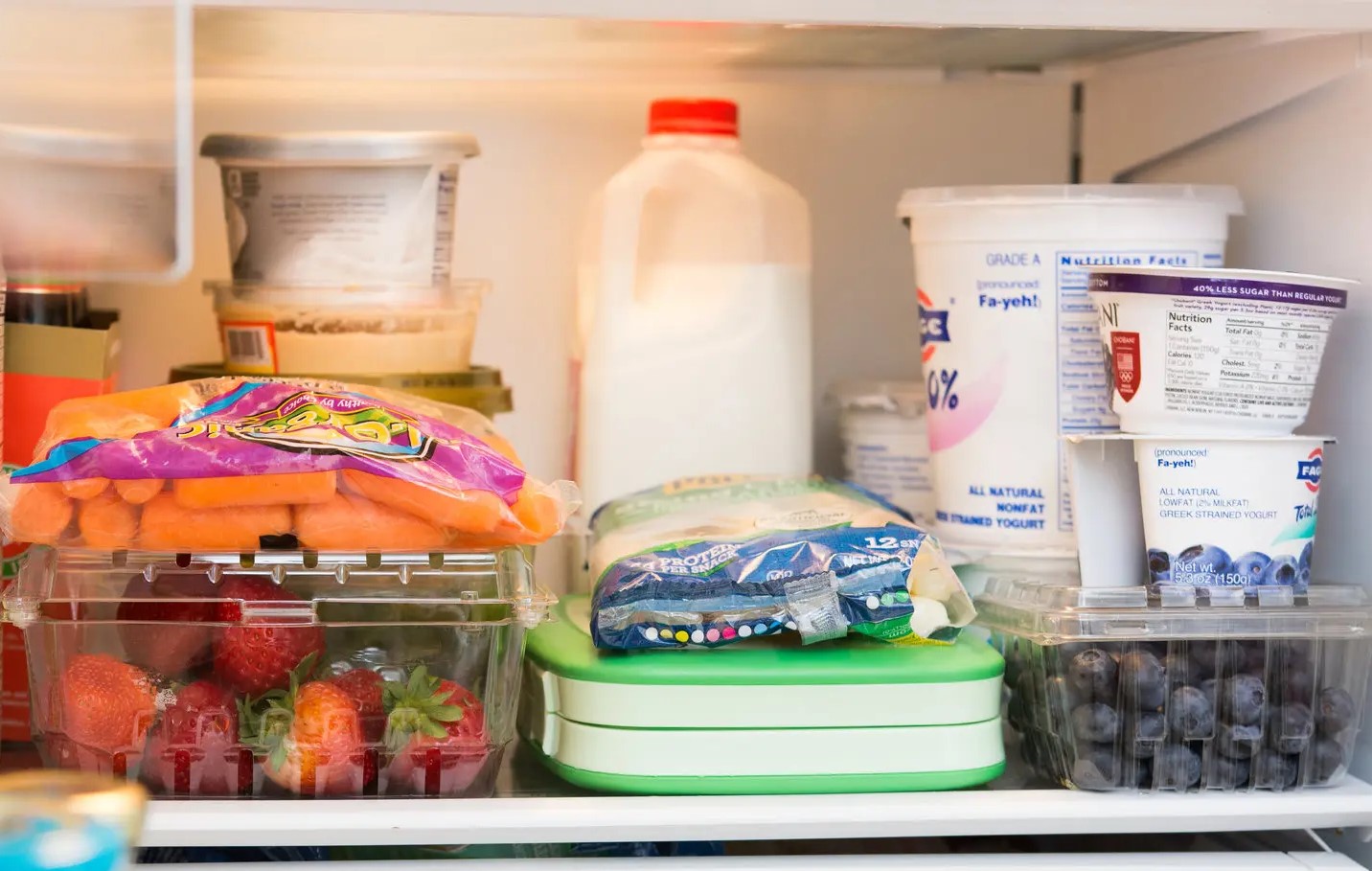}}}                  \\ 

\midrule[2pt]
\multicolumn{1}{c}{User}      & What are the meals that I can cook with these? \\ \midrule[2pt]
\multicolumn{1}{c}{MM-LIMA} & There are various meals that can be cooked with the ingredients shown in the refrigerator. Here are some ideas:

* Yogurt parfait: Place yogurt in a bowl and top with fruit, granola, or nuts.

* Fruity smoothies: Combine yogurt with frozen fruit and blend until smooth. Add honey or sugar to taste.

* Cottage cheese and fruit: Serve cottage cheese with fresh fruit or berries.

* Breakfast smoothie: Combine yogurt, milk, fruit, and protein powder in a blender.

* Grilled cheese sandwich: Spread butter on two slices of bread and add cheese. Grill until golden brown.

* Oatmeal and fruit: Cook oatmeal according to package directions and top with fruit and honey or sugar.

* Spinach and cheese quiche: Mix spinach with cheese and eggs, then pour into a pie crust. Bake until golden brown.

These are just a few ideas to get started. You can also mix and match ingredients to create your own recipes.
 \\ \midrule
\multicolumn{1}{c}{MiniGPT-4} & This refrigerator contains various types of food, including milk, yogurt, cereal, fruits, and vegetables. \\ \midrule[2pt]

\multicolumn{1}{c}{User}      & I would like to do the fruit salad. Show me the recipe. \\ \midrule[2pt]
\multicolumn{1}{c}{MM-LIMA} & Sure, here’s a recipe for a simple fruit salad:

Ingredients:

* 1 kg of mixed fruit (such as apple, banana, grapes, pineapple, and watermelon)

* 1 cup of yogurt

* 1 tablespoon of honey

* 1 tablespoon of lime juice

* 1/4 cup of raisins

Instructions:

1. Wash and slice the fruit.
2. In a large bowl, mix the yogurt, honey, and lime juice.
3. Add the fruit to the bowl and mix everything together until the fruit is evenly coated in the yogurt mixture.
4. Add the raisins to the bowl and stir.
5. Serve the fruit salad chilled.

You can adjust the amount of fruit and yogurt to your liking. You can also add other ingredients such as granola, nuts, or even chia seeds for added texture and nutrition. Enjoy!
 \\ \midrule
\multicolumn{1}{c}{MiniGPT-4} & Sure, here’s a recipe for a simple fruit salad:

Ingredients:

* Bananas, peeled and sliced

* Apples, peeled and sliced

* Grapes

* Strawberries

* Blueberries

* Rice

* Sugar

* Lemon juice

Instructions:

1. In a large bowl, mix together the sliced bananas, apples, grapes, strawberries, blueberries, and rice.
2. In a separate bowl, mix together the sugar and lemon juice until the sugar is fully dissolved.
3. Pour the sugar mixture over the fruit mixture and toss to coat everything evenly.
4. Serve the fruit salad cold or chilled, garnished with lemon wedges or mint leaves if desired.

That’s it! This simple fruit salad is a great way to get in some healthy fruits and veggies while satisfying your sweet tooth. Enjoy! \\ \bottomrule[2pt]
\end{tabular}}
\caption{Generating chain-of-thought responses.} \label{tab:cot}
\end{table}

\section{Selected Samples Versus Original Samples} \label{samples}
\begin{figure}[t]
\centering
  \includegraphics[width=0.95\textwidth]{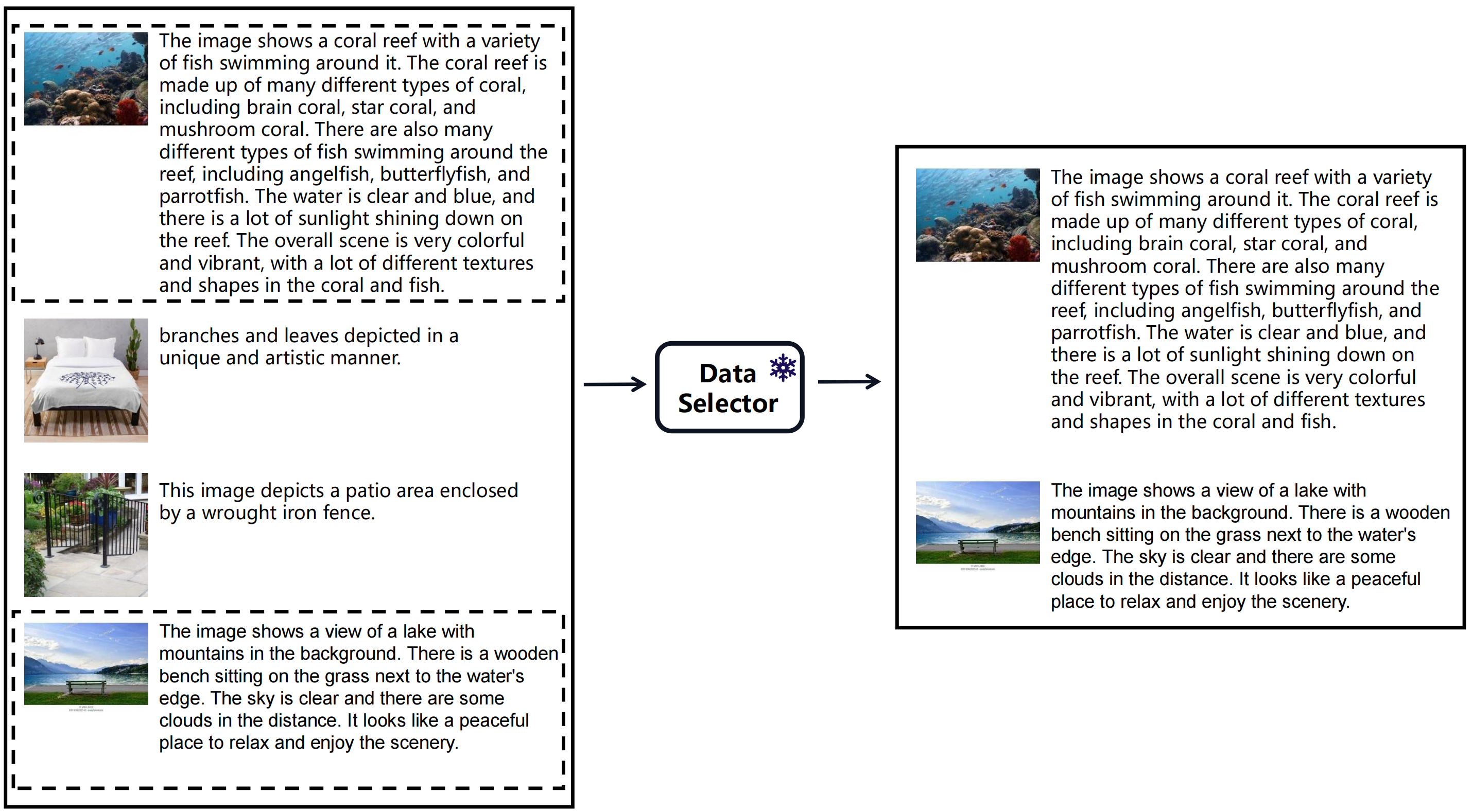} 
  \caption{Data selector can filter out low-quality data (e.g., inappropriate grammar and incomplete expressions).} 
  \label{fig:show} 
\end{figure}

We compare the selected 200 data points with the original dataset and find that the selected 200 data points exhibit higher quality. This is reflected in the more comprehensive, detailed, and error-free language representation of the data, with no apparent grammatical or syntactical errors. It demonstrates that the data selector can effectively choose data aligned with human preferences in Figure~\ref{fig:show}. 

\section{Limitation and Future Works}
While achieving promising performance, the proposed framework also has some inherent limitations.
In our experiments, we only explore the correlation between indicators and genuine quality labels on fixed {MiniGPT-4-7B}~\cite{zhu2023minigpt}.
In forthcoming research, extending the analysis to include other models of varying sizes, such as MiniGPT-4-13B and LLaVA~\cite{liu2023llava}, could provide additional insights. Besides, developing a more generalized version of the data selector, such as considering more indicators for evaluation, is another potential avenue for exploration. 
Additionally, undertaking multimodal instruction mining
may provide further insights and unveil new opportunities in this domain. This could enable a broader and more comprehensive understanding of this field.